\documentclass[manuscript,screen]{acmart}

\usepackage{graphicx}
\usepackage{booktabs}
\usepackage{tabu}

\AtBeginDocument{%
  }

\setcopyright{acmlicensed}
\copyrightyear{2025}
\acmYear{2025}
\acmDOI{XXXXXXX.XXXXXXX}

\begin{document}

\title[Better Bill GPT]{Better Bill GPT: Comparing Large Language Models against Legal Invoice Reviewers}


\author{Nick Whitehouse, Nicole Lincoln, Stephanie Yiu, Lizzie Catterson, Rivindu Perera}
\email{nick.whitehouse@onit.com}
\affiliation{%
  \institution{AI Center of Excellence, Onit Inc}
  \city{Aukland}
  \country{New Zealand}
}

\renewcommand{\shortauthors}{Whitehouse et al.}

\begin{abstract}
Legal invoice review is a costly, inconsistent, and time-consuming process, traditionally performed by Legal Operations, Lawyers or Billing Specialists who scrutinise billing compliance line by line. This study presents the first empirical comparison of Large Language Models (LLMs) against human invoice reviewers — Early-Career Lawyers, Experienced Lawyers, and Legal Operations Professionals—assessing their accuracy, speed, and cost-effectiveness. Benchmarking state-of-the-art LLMs against a ground truth set by expert legal professionals, our empirically substantiated findings reveal that LLMs decisively outperform humans across every metric. In invoice approval decisions, LLMs achieve up to 92\% accuracy, surpassing the 72\% ceiling set by experienced lawyers. On a granular level, LLMs dominate line-item classification, with top models reaching F-scores of 81\%, compared to just 43\% for the best-performing human group. Speed comparisons are even more striking — while lawyers take 194 to 316 seconds per invoice, LLMs are capable of completing reviews in as fast as 3.6 seconds. And cost? AI slashes review expenses by 99.97\%, reducing invoice processing costs from an average of \$4.27 per invoice for human invoice reviewers to mere cents. These results highlight the evolving role of AI in legal spend management. As law firms and corporate legal departments struggle with inefficiencies, this study signals a seismic shift: The era of LLM-powered legal spend management is not on the horizon, it has arrived. The challenge ahead is not whether AI can perform as well as human reviewers, but how legal teams will strategically incorporate it, balancing automation with human discretion.
\end{abstract}




\maketitle

\section{Introduction}

Our previous study, \textit{Better Call GPT} \cite{martin2024better}, demonstrated that Large Language Models (LLMs) \cite{vaswani2017attention} can achieve human-comparable accuracy in contract review tasks while significantly reducing time and costs. By establishing a foundational benchmark for the evaluation of legal context, the aforementioned research highlighted the potential for LLMs to improve efficiency in contract review. Building on these findings, the present study examines the application of LLMs in legal invoice processing, investigating their ability to match or surpass human expertise in identifying non-compliant line items, enforcing billing policies, and resolving disputes.

Legal invoice processing is a critical, yet resource-intensive aspect of legal operations that requires meticulous scrutiny to ensure compliance with billing guidelines, letters of engagement, agreements (such as alternative fee arrangements), and regulatory requirements. Traditional invoice review processes are labour intensive, time-consuming, and prone to inconsistencies introduced in the reasoning process for disputes. The introduction of LLMs into this domain presents a compelling opportunity to automate and standardise invoice review. However, the effectiveness of LLMs in this context remains an open question which requires a systematic evaluation of their capabilities, limitations, and comparative advantages over human professionals.

To address these gaps in current research, we formulate three research questions to assess the accuracy, efficiency, and cost-effectiveness of LLMs in legal invoice processing:

\vspace{2mm} 
\noindent \textbf{To what extent can LLMs accurately identify non-compliant line items in legal invoices compared to human invoice reviewers?}
\vspace{1mm} 

\begin{quote}
This question examines the ability of LLMs to detect billing violations, assess compliance with billing guidelines, and ensure accuracy in legal invoicing relative to human invoice reviewers.
\end{quote}

\vspace{2mm} 
\noindent \textbf{How does the efficiency of LLMs in processing legal invoices compare to that of human invoice reviewers?}
\vspace{1mm} 
\begin{quote}
Here, we explore whether LLMs can analyse invoices, enforce billing policies, and make decisions on dispute or approval more quickly than the traditional human-led review process.
\end{quote}

\vspace{2mm} 
\noindent \textbf{What is the comparative cost-effectiveness of LLM-driven legal invoice review relative to the traditional human-led review process?}
\vspace{1mm} 
\begin{quote}
The focus is on determining whether LLMs offer a financially viable alternative by reducing operational costs while maintaining accuracy and compliance.
\end{quote}

\vspace{2mm} 
By systematically addressing these research questions, this study aims to provide a comprehensive evaluation of LLMs in legal invoice processing, offering insights into their accuracy, efficiency, and cost-effectiveness. Beyond assessing their performance, this research seeks to establish a benchmark for legal invoice review, setting a standardised framework against which future AI-driven solutions can be measured. Such a benchmark has the potential to advance the industry by providing legal practitioners, technology providers, and policymakers with a reliable point of reference for evaluating and adopting AI-assisted invoice review methodologies. Through this approach, this study contributes to the broader discourse on AI's role in legal operations, fostering innovation while ensuring adherence to regulatory standards and guidelines.

\section{Related Work}

The automation of invoice processing \cite{hedberg2020automated} has progressed from conventional rule-based extraction techniques and structured data capture toward more advanced methodologies, including machine learning-driven anomaly detection. Most recently, advancements in LLMs have introduced a new paradigm shift in financial reasoning. While early AI-driven solutions enhanced operational efficiency by automating routine data extraction, their limitations became evident in more complex tasks. These systems struggled to interpret nuanced billing justifications, enforce intricate compliance policies, and dynamically mediate disputes. The integration of LLMs marks a pivotal shift, offering the potential to bridge these gaps by contextualising financial data, reasoning through justifications, and adapting to regulatory frameworks in real time.

However, despite the advancements in invoice automation, the existing literature exhibits a notable gap in research dedicated to LLM-based invoice processing. While LLMs have demonstrated significant potential in various domains, their application in financial document automation remains unexplored. Moreover, the application of LLM-based invoice processing is even more unexplored within the legal domain. Given the complexity of legal billing structures, regulatory compliance requirements, and the need for precise justification of charges, existing automation techniques often fall short of addressing the unique challenges inherent in legal invoice processing. As a result, this section broadens the scope of the review of early literature by incorporating traditional machine learning-based techniques that have historically supported invoice processing advancements.

Anchoori \cite{Anchoori2024AIDrivenDP} presents an AI-driven invoice processing framework that integrates Optical Character Recognition (OCR) with machine learning to automate data extraction, achieving a 97.8\% accuracy rate. The study highlights the efficiency gains from automated invoice validation which significantly reduces the processing time and manual errors. However, despite its success in structured data extraction, this approach does not engage in areas that require interpretive decision-making beyond predefined rules such as compliance reasoning or invoice dispute resolution. This is noted as the major gap in this research where such areas are key for automating the invoice processing workflows. In contrast, our study focuses on evaluating whether LLMs can analyse ambiguous invoice descriptions, apply contractual billing guidelines, and determine approval/dispute decisions, which replicate expert-level judgment rather than simple validation.

Similarly, Constantinou and Kabiri \cite{constantinou2020detecting} investigate machine learning-based anomaly detection in legal invoices by employing Support Vector Machines (SVMs) to flag irregular billing patterns. Their study demonstrates that machine learning can effectively identify overcharges, duplicate invoices, and non-standard fee structures, providing an essential tool for financial oversight. However, their approach remains limited to statistical anomaly detection, without assessing the legitimacy of charges based on billing justifications or legal compliance requirements. Our research extends beyond anomaly detection by testing whether LLMs can reason through justifications, enforce compliance policies, and dynamically approve or dispute based on legal and financial guidelines.

Building on these developments, Seara \cite{searainvoice} introduces process mining and machine learning models to invoice approval workflows to predict final invoice statuses. By combining transition systems with predictive analytics, this study successfully forecasts invoice completion times and outcomes, resulting in more efficient resource allocation within finance departments. However, this approach primarily focuses on workflow efficiency optimisation rather than deep invoice compliance reasoning. While the study offers a data-driven approach to process flow analysis, it does not examine whether AI can replace or augment human expertise in evaluating invoice compliance and dispute resolution. Our research differs by benchmarking LLMs against human invoice reviewers in real-world invoice compliance tasks, testing their ability to enforce billing guidelines and make compliance decisions which are traditionally handled by financial professionals.

Although with advances in data extraction, anomaly detection, and process mining, existing AI-driven invoice automation approaches remain constrained by their reliance on structured inputs and predefined models. They do not engage in high-order interpretive reasoning, contractual enforcements (through billing guidelines), or real-time judgments. These domains, however, represent a set of key domains where LLMs present a fundamental shift in AI capabilities. Unlike previous studies, which focus on statistical patterns and efficiency optimisation, our research investigates whether LLMs can match or exceed human-level accuracy in invoice compliance enforcement. By directly comparing LLM performance against Early-Career and Experienced human invoice reviewers, we introduce an industry-specific benchmark evaluating AI's role in decision-making, compliance enforcement and financial oversight within invoice auditing workflows in the legal domain.

\section{Methodology}

The following sections outline the research methodology used for comparing human invoice reviewers against LLMs in legal invoice processing tasks, including data collection, analysis procedures, and model selection criteria. 

\subsection{Overall Design and Objectives}

This study was designed to compare how Experienced and Early-Career human invoice reviewers perform relative to LLMs when identifying non-compliant line items and determining whether to approve or dispute invoices.  

\subsection{Ethics Policy}
This research adhered to the ethical standards defined by Onit Inc., covering data collection, analysis, and participation in primary and secondary research. 

All participants were fully informed about the research purpose, data usage, and their rights, including the right to withdraw at any time. Any identifiable information in the collected data was replaced to ensure participant anonymity.

The dataset comprised sanitised client invoices supplemented by synthetic data. All client data usage required explicit consent and underwent anonymisation to remove Personally Identifiable Information (PII) in accordance with input control requirements.

\subsection{Data Collection and Analysis}

This study involved two distinct groups: the Ground Truth Team and the Human Invoice Review Team. The Ground Truth Team was responsible for establishing whether invoices and line items should be disputed or approved.  Their decisions on whether invoices and line items should be disputed or approved served as the benchmark for comparison.  The Human Invoice Review Team, composed of Experienced and Early-Career invoice reviewers, independently assessed the same invoices. Their performance was evaluated relative to the established ground truth and used as the human baseline for comparison with LLMs. 

The data set used for this study comprised 50 legal invoices, including anonymised client invoices and realistic synthetic examples. To prevent overfitting, a balanced dataset of real and synthetic invoices was ensured, incorporating natural variability in billing styles, including formatting and language. All synthetic invoices underwent human validation against real-world billing patterns through manual inspection by a legal professional. These invoices covered multiple billing scenarios, including various practice areas, billing rates, and levels of compliance violation complexity.

In total, the data set contained 492 line items encompassing a broad range of expense categories in different formatting styles. These characteristics ensured that the dataset accurately represented real-world invoice review challenges.

Invoice compliance was evaluated using a standardised set of legal billing guidelines adapted from widely recognised best practices. Key areas of assessment include timely submission, ensuring invoices are submitted within the required time frame; block billing, where multiple unrelated tasks in a single entry lack itemisation; unauthorised tasks, identifying non-billable activities such as clerical work or excessive research; expense caps, verifying that charges do not exceed pre-defined limits for specific cost categories; and timekeeper limits, ensuring that billed hours per individual comply with contractual or industry standards.

\subsubsection{Ground Truth Review}

To establish the ground truth for invoice compliance, we assembled a panel of nine experienced invoice reviewers, including legal bill review specialists, current and former legal professionals from law firms, and corporate legal department staff. All panelists had substantial experience in invoice review, ensuring that the ground truth reflected expert-level decision-making and created a high-quality benchmark for evaluating AI and human performance. This panel was structured to align with real-world industry practices, ensuring that the established ground truth was rigorous and representative of standard legal billing review methodologies. 

\subsubsection{Human Invoice Review}

The Human Invoice Review group was used to establish human data against which LLMs were compared for accuracy and efficiency.   

This group included practising lawyers (from law firms or in-house roles), legal operations professionals, and finance professionals who assist with billing oversight. Post-Qualification Experience (PQE) \cite{kirton2020solicitors} was used to serve as a reasonable proxy for dedicated invoice review experience. PQE is a widely recognised metric in the legal profession used to assess lawyer seniority, competency, and role progression, providing a standardised measure of a lawyer's experience post-qualification. As lawyers progress in their careers, increased PQE typically corresponds with greater responsibilities, including financial management and invoice review. 

Participants were classified as Experienced Lawyer, Early-Career Lawyer or Experienced Non-Lawyer based on their legal training and practical invoice review experience, ensuring a realistic distribution of expertise levels that reflects real-world legal invoice compliance processes.

The classification of reviewers who meet the threshold of being ‘experienced’ follows industry norms, as reflected in common job descriptions in the US legal and financial sectors. Lawyers with five or more years of PQE are often seen as experienced senior associates, capable of handling significant financial oversight, with responsibilities such as managing case costs and invoice disputes, supervising invoices, and ensuring billing compliance. For non-lawyers, legal operations and finance professionals develop a comparable level of significant specialist-level invoice compliance expertise after approximately three years of dedicated invoice review experience, making this an appropriate equivalent.  

The categories within this group are as follows: 
\begin{enumerate}
    \item Early-Career Lawyer - Attorneys or former practising lawyers (law firm or in-house) with less than five years of PQE, likely having limited direct exposure to billing and invoice review processes.
    \item Experienced Lawyer - Attorneys or former practising lawyers (law firm or in-house) with five or more years of PQE, expected to have substantial experience reviewing legal invoices and applying billing compliance standards.
    \item Experienced Non-Lawyer: Legal operations or finance professionals with at least three years of direct involvement in invoice compliance (e.g., reviewing, approving, or disputing invoices).
\end{enumerate}

\subsection{Experimental Procedure}

This study followed a structured experimental procedure to ensure a systematic and unbiased comparison of legal invoice review performance across human invoice reviewers and LLMs.

\subsubsection{Task Description}
Both the Ground Truth Team and the Human Invoice Review Team completed the same set of tasks under standardised conditions to ensure comparability in their assessments. Each group completed the following: 
\begin{enumerate}
    \item Reviewed a set of invoices, which included both anonymised client invoices and synthetic examples. Each member of the Ground Truth Team evaluated the full set of invoices, while members of the Human Invoice Review Team reviewed a randomly assigned subset of invoices.  
    \item Analysed each invoice entry for potential violations of the billing guidelines, flagging line items that do not comply.
    \item  For each disputed line item, provided reasoning for the type of guideline violation, such as block billing, unauthorised tasks, expense cap violations, or timekeeper limit breaches.
    \item Based on their compliance assessment, determine whether each invoice should be approved or disputed.
    \item Determine whether they would actually dispute the invoice in practice, considering pragmatic factors such as client relationships, precedent, business impact, or practical enforceability of billing guidelines.
\end{enumerate}

All reviews were independent. Human invoice reviews were single-pass with no revisions allowed, and LLMs processed invoices sequentially, preventing exposure to prior outputs. 

\subsubsection{Instructions}
Both human invoice reviewers and LLMs were provided with a standardised set of instructions detailing how to review invoices, identify violations, and explain non-compliant items. 

Human invoice reviewers were provided with a standardised, written set of guidelines outlining billing compliance rules, common violations, and documentation requirements. Structured spreadsheets were also provided to each human invoice reviewer to document violations and record justifications for approval/dispute decisions. 

LLMs received structured input prompts that mirror the instructions provided to human invoice reviewers. The prompts explicitly instructed the models to:
\begin{itemize}
    \item Analyse the invoice line by line for compliance violations.
    \item Explain violations according to specific guideline rules.
    \item Make a final determination on approval or dispute, including a justification based on billing guidelines.
\end{itemize}

Human invoice reviewers were also asked to record the amount of time spent reviewing each invoices, and LLM processing time was manually or automatically logged. 

\subsection{Implementation and Technical Details}
\subsubsection{AI Model Selection}

This study included six LLMs (OpenAI o1, GPT-4o, Claude 3.5 Sonnet, Claude 3.7 Sonnet with extended thinking, Gemini 2.0 Flash Thinking Experimental 01-21, and DeepSeek R1) selected based on their state-of-the-art performance, widespread adoption, and general-purpose capabilities in legal text processing. While the study does not include specialised legal AI products, this decision is intentional for the following reasons: 
\begin{enumerate}
    \item Benchmarking General Models: Legal invoice compliance is an applied task, not solely a legal reasoning challenge. General LLMs provide a baseline against which future legal-specific fine-tuned models can be compared.
    \item Real-World Adoption: Many legal teams integrate general LLMs into workflows rather than relying solely on domain-specific AI due to cost, ease of implementation, and API-based automation potential.
    \item Reproducibility \& Transparency: Legal AI models often operate as black-box systems, whereas general LLMs provide structured, explainable outputs.
\end{enumerate}

The selected models varied in architecture, scale, and provider specialisation, ensuring diversity in the comparison. Each model was evaluated under identical conditions, with standardised prompts and task descriptions, ensuring a fair and unbiased comparison. 

DeepSeek was hosted internally on AWS, allowing for controlled inference and ensuring data privacy during invoice processing. All other models were accessed via their respective API endpoints under standard usage conditions. 

\subsubsection{Prompt Engineering}
Each LLM required tailored prompts to optimise performance. The prompts each included the following elements: 
\begin{itemize}
    \item A role definition instructing the LLM to act as a legal invoice reviewer.
    \item Task instructions specifying compliance checks against billing guidelines to identify line items which violate guidelines and make approval/dispute determinations for each invoice.
    \item Contextual information about the invoice, billing rules, and organisational policies which is typically provided to invoice reviewers.
\end{itemize}

Prompts and model-specific refinements were tested iteratively to improve accuracy and reliability in dispute identification. 

\subsection{Metrics and Evaluation Criteria}
\subsubsection{Accuracy Measures}

To evaluate the accuracy of invoice compliance assessments, we used the Ground Truth Reviewer's consensus as the basis of correctness. Each response was scored using standard classification metrics: 
\begin{itemize}
    \item Precision (Positive Predictive Value) – The proportion of correctly identified non-compliant line items relative to all flagged violations.
    \item Recall (Sensitivity) – The proportion of actual non-compliant line items correctly identified.
    \item F-Score – The harmonic mean of precision and recall, providing a balanced measure of accuracy.
\end{itemize}
Each reviewer type (Experienced Lawyer, Experience Non-Lawyer, Early-Career Lawyer, and LLM) was assessed based on how well their approval/dispute decisions aligned with the ground truth, and how frequently they correctly identified line item violations. 

\subsubsection{Time Efficiency}
To measure time efficiency, we calculated the average time per invoice for each reviewer group by aggregating individual per-invoice review times within each category. This allowed for a comparative analysis of how efficiency varied across different levels of human expertise and LLM performance. 

For LLMs, time efficiency was assessed using the Average Processing Time per Invoice, where Processing Time was measured as the time elapsed between submitting a compliance check invoice and prompt, and receiving the model’s output.  

\subsubsection{Cost Analysis}

The cost analysis considered both human invoice review rates and AI processing costs, accounting for scalability and cost-performance trade-offs to provide a range of acceptance for infrastructure costs. The cost comparison assessed whether AI-driven legal invoice review can provide a cost-effective alternative to human invoice review, while maintaining accuracy and compliance standards.  All amounts in this subsection are in USD. 

The hourly rates for human invoice reviewers were determined using industry compensation surveys and salary data aggregators, such as RobertHalf's 2025 Salary Guide \cite{roberthalf2025salaryguide}. Early-Career Lawyers with less than five years of experience had an estimated average hourly rate of \$61, while those with five or more years of experience averaged \$79 per hour. Non-attorney reviewers, including legal billing specialists and legal operations professionals with more than three years of experience, had an estimated average hourly rate of \$47 per hour. 

The AI cost evaluation includes API-based commercial models (OpenAI o1, GPT-4o, Claude 3.5 Sonnet, Claude 3.7 Sonnet, and Gemini 2.0 Flash Thinking Experimental 01-21) and self-hosted models (DeepSeek R1 on AWS EC2 GPU instances). AI costs vary depending on deployment method, processing scale, and legal invoice complexity. 

\section{Evaluation and Results}

\subsection{Ground Truth Invoice Reviewer Comparison }
\label{human-reviewer-comparison}

We assessed the reliability of the ground truth labels used for both invoice-level and line-item-level decisions. Cronbach's Alpha was used to measure the internal consistency of the Ground Truth Team's agreement on invoice-level and line-item-level classifications. Fleiss' kappa was used to assess the inter-rater agreement for line item dispute reasoning. 
\begin{itemize}
    \item Invoice Approval/Dispute Decision: 0.868 Cronbach's Alpha, indicating strong agreement among human invoice reviewers. 
    \item Line Item Approval/Dispute Decision: 0.832 Cronbach's Alpha, reflecting a high level of consistency. 
    \item  Line Item Dispute Reasoning: 0.331 Fleiss' kappa, suggests a fair inter-rater agreement on the reasons for disputing individual line items. 
\end{itemize}

While there was a strong agreement with regard to the approval/dispute decisions at both the invoice and line item level, the lower Fleiss' kappa score for line item dispute reasons indicates subjectivity plays a significant role in determining why a line item is disputed. 

\subsection{AI vs. Human Performance}

To comprehensively evaluate the effectiveness of LLMs and human invoice reviewers, we assessed performance at both the invoice level and the line item level. By analysing both decision types, we aimed to understand not only whether LLMs and human invoice reviewers correctly classify invoices as a whole, but also whether they effectively assess the legitimacy of individual charges. These two levels of evaluation reflect the realities of legal invoice review, where both broad and detailed assessments impact financial and compliance decisions. 

\subsubsection{Invoice Approve/Dispute Decision Accuracy}
Table\ref{tab:dispute} shows that LLMs generally outperformed human invoice reviewers in determining whether an invoice should be approved or disputed. GPT-4o and Gemini 2.0 Flash Thinking achieved the highest accuracy, both with an F-score of 0.920. There was a notable drop in accuracy among the next highest scoring group of LLMs, with Claude 3.7 Sonnet, Claude 3.5 Sonnet, and o1 each scoring 0.840. DeepSeek R1 trailed slightly behind with an F-score of 0.820. 

By comparison, the top performing human invoice reviewers, Experienced Lawyers, achieved an F-score of 0.720, significantly below even the lowest scoring LLMs. Early-Career Lawyers and Experienced Non-Lawyers were slightly lower, scoring 0.700 and 0.680, respectively.  These results indicate that LLMs demonstrate greater consistency and reliability in invoice-level decisions compared to human invoice reviewers.



\begin{table}[ht]
    \begin{center}
        \begin{tabu}{X[2,l] X[1,l]}
            \toprule
         \textbf{Role}& \textbf{F-score}\\
    \midrule
         Gemini 2.0 Flash Thinking
& 0.9200\\
         GPT-4o
& 0.9200\\
         Claude 3.7 Sonnet
& 0.8400\\
 o1
&0.8400\\
         Claude 3.5 Sonnet
& 0.8400\\
 DeepSeek R1
&0.8200\\
         Experienced Lawyers
& 0.7206\\
 Early Career Lawyers
&0.7024\\
 Experienced Non-Lawyers&0.6754\\
 \bottomrule
            
        \end{tabu} 
    \end{center}
    \caption{Performance ranked by accuracy score for invoice decision}
    \label{tab:dispute}
\end{table}

\subsubsection{Line Item Classification F-Score}
Table \ref{tab:line items} summarizes the results of the line-item classification task, where LLMs also emerged as the top performers. The highest F-score was achieved by Gemini 2.0 Flash Thinking, at 0.806. This was followed by Claude 3.7 Sonnet at 0.643 and GPT-4o at 0.631. The lowest scoring LLM, Claude 3.5 Sonnet, scored 0.522. In contrast, the top performing human invoice reviewer group, Experienced Lawyers, had an F-score of 0.429, and Experienced Non-Lawyers scored the lowest of all reviewers, at 0.368.

\begin{table}[ht]
    \begin{center}
        \begin{tabu}{X[2,l] X[1,l]}
            \toprule
         \textbf{Role}& \textbf{F-score}\\
    \midrule
 Gemini 2.0 Flash Thinking
&0.8058\\ 
 Claude 3.7 Sonnet
&0.6426\\ 
 GPT-4o
&0.6307\\ 
 o1
&0.5939\\
 DeepSeek R1
&0.5938\\
 Claude 3.5 Sonnet
&0.5218\\
 Experienced Lawyers
&0.4292\\
 Early Career Lawyers&0.3982\\ 
 Experienced Non-Lawyers&0.3677\\
 \bottomrule
            
        \end{tabu} 
    \end{center}
    \caption{Performance ranked by F-score for line item decision}
    \label{tab:line items}
\end{table}

While LLMs also outperformed human invoice reviewers in line item classification, the results should be considered in light of the variability of the ground truth reasons for dispute. As noted above, the dispute rationales showed a low agreement, suggesting subjectivity in how individual line items were assessed. This may explain some inconsistencies in both the baseline human and LLM performance on line item classification.

\subsection{Time Efficiency}
A time analysis was conducted to compare the efficiency of human invoice reviewers and LLMs in reviewing legal invoices. 

The results, summarised in Table \ref{tab:Time}, revealed notable differences across groups. Among the human invoice reviewers, Experienced Lawyers were the most efficient, averaging 194.75 seconds per invoice. Early-Career Lawyers took longer at 257.5 seconds, while Experienced Non-Lawyers had the slowest review times, averaging 316.93 seconds per invoice. 

In contrast, all LLMs demonstrated significantly higher time efficiency, completing reviews in mere seconds. The slowest LLM, DeepSeek, took 21.48 seconds, while GPT4o was the fastest at just 3.61 seconds.  These findings indicate that LLMs significantly outperform human invoice reviewers in time efficiency when analysing legal invoices.

\begin{table}[ht]
    \begin{center}
        \begin{tabu}{X[2,l] X[1,l]}
            \toprule
         \textbf{Role}& \textbf{Average Time per Invoice (Seconds)}\\ 
    \midrule
 GPT-4o
&3.61
\\ 
 Claude 3.5 Sonnet
&6.34
\\ 
 Gemini 2.0 Flash Thinking
&8.68
\\ 
 Claude 3.7 Sonnet
&17.33
\\ 
 o1
&17.87
\\ 
 DeepSeek R1
&21.48
\\ 
 Experienced Lawyers
&194.75
\\
 Early Career Lawyers&257.5
\\ 
 Experienced Non-Lawyers&316.93\\ 
 \bottomrule
            
        \end{tabu} 
    \end{center}
    \caption{Time Comparison}
    \label{tab:Time}
\end{table}

\subsection{Cost Comparison }
A cost analysis was also conducted to compare the expenses associated with human invoice reviewers and LLMs in reviewing legal invoices. 

As shown in Table \ref{tab:Cost}, human invoice reviewers incurred significantly higher costs per invoice. Experienced Non-Lawyers were the most cost-effective of the human invoice reviewers, averaging \$4.17 per invoice. Experienced Lawyers and Early Career Lawyers were both slightly more expensive, averaging \$4.27 and \$4.47 per invoice, respectively.  

Conversely, LLMs offered dramatically lower costs per invoice. Gemini (currently free as a preview model), GPT-4o, Claude 3.5 Sonnet, and DeepSeek all operated under or at \$0.01 per invoice. OpenAI was the most expensive LLM, costing \$0.13 per invoice. These findings highlight that LLMs provide a cost-effective alternative to human invoice reviewers, reducing expenses by over 99\%.

\begin{table}[ht]
    \begin{center}
        \begin{tabu}{X[2,l] X[1,l]}
 \toprule
         \textbf{Role}& \textbf{Average Cost per Invoice (USD)}\\
    \midrule
         Gemini 2.0 Flash Thinking
& \$0.00\\
 GPT-4o
& \$0.01\\
         Claude 3.5 Sonnet
& \$0.01\\
         DeepSeek R1
& \$0.01\\
         Claude 3.7 Sonnet
& \$0.02\\
         o1
& \$0.13\\
         Experienced Non-Lawyers& \$4.17\\
         Experienced Lawyers
& \$4.27\\
         Early Career Lawyers& \$4.47\\
     \bottomrule

        \end{tabu} 
    \end{center}
    \caption{Cost Comparison}
    \label{tab:Cost}
\end{table}

\section{Subjectivity Analysis}

As mentioned in Section \ref{human-reviewer-comparison}, the presence of subjectivity became evident during the analysis of inter-rater agreement, as reflected in a Fleiss' kappa value of 0.331 for line item level reasons leading to disputes. This relatively low agreement suggests variability in reviewers' assessments, leading to a deeper investigation into the effect of subjectivity among legal invoice reviewers. We examined the extent to which billing guideline enforcement fosters objective decision-making, as opposed to the discretionary judgment exercised by individual reviewers. Understanding this relationship is essential to assess the consistency and reliability of the legal invoice review process.

\begin{table}[ht]
    \begin{center}
        \begin{tabu}{X[1,l] X[2,l] X[1,r] X[1,r] X[1,r] X[1,r]}
            \toprule
             \textbf{Reviewer} &       \textbf{Experience Level}  &  \textbf{Decisions overruled} &  \textbf{Subjectivity} &  \textbf{\ Objective Match / GT \%} &  \textbf{Subjective Match / GT \%}\\
             \midrule
           10-R &     Experienced Lawyer &                                    0 &          0 &          71 &            71 \\
    11-R &    Early-Career Lawyer &                                       7 &          33 &          71 &            86 \\
    12-R &    Early-Career Lawyer &                                    5 &          24 &          71 &            76 \\
    13-R &     Experienced Lawyer &                                         9 &          43 &          67 &            62 \\
    14-R & Experienced Non-Lawyer &                                      8 &          38 &          48 &            86 \\
    15-R &    Early-Career Lawyer &                                     3 &          14 &          67 &            81 \\
    16-R &    Early-Career Lawyer &                                4 &          19 &          76 &            86 \\
    17-R &     Experienced Lawyer &                                   4 &          19 &          81 &            71 \\
    18-R &     Experienced Lawyer &                            1 &          5 &          86 &            81 \\
    20-R & Experienced Non-Lawyer &                                    0 &          0 &          70 &            70 \\
    21-R & Experienced Non-Lawyer &                                     5 &          25 &          85 &            90 \\
    \midrule
     AVG &                     &                                   4 &          20 &          72 &            78 \\

         \bottomrule
            
        \end{tabu} 
    \end{center}
    \caption{Subjectivity Analysis}
    \label{tab:subjectivity-analysis}
\end{table}

To examine the extent of subjectivity, we analysed the decision-making patterns of human invoice reviewers - a group comprised of varying levels of experience. Table \ref{tab:subjectivity-analysis} presents key metrics, including the number of approved and disputed invoices, instances where decisions are overruled, and the impact of subjective judgement on the accuracy of the decision.

Our findings reveal notable variations in human invoice reviewer behaviour. Experienced Lawyers exhibited a relatively lower rate of decision overruling, indicating strong adherence to billing guidelines. In contrast, Early-Career Lawyers and Experienced Non-Lawyers demonstrated a higher tendency to overrule their decisions when they make subjective judgements.  Based on the results, it is clear that, on average, 20\% of human invoice reviewers' decisions exhibit subjectivity where they would agree to deviate from strict adherence to billing guidelines in favour of discretionary judgements. Interestingly, our analysis also uncovers that by incorporating subjective decision-making, alignment with ground truth can be increased by 6\% on average. This suggests that a degree of discretion allows human invoice reviewers to make more contextually appropriate decisions, aligning more closely with the expected outcomes of the review process.

However, these findings also raise important questions about consistency and reliability. Although subjectivity can improve decision accuracy in certain cases, as noted above, it also introduces variability between human invoice reviewers, which can potentially lead to inconsistencies in dispute resolution in the legal spend management domain. This emphasises the need for a structured yet adaptable approach that evaluates billing guidelines while ensuring objective knowledge, mirroring a consistent and knowledge-infused workflow that can be expected from LLMs.

\section{Discussion and Further Work}

This study set out to answer three core questions regarding the viability of LLMs in legal invoice review: (1) Can LLMs accurately identify non-compliant line items compared to human invoice reviewers? (2) Are LLMs more efficient in reviewing invoices? And (3) Do LLMs offer a cost-effective alternative to traditional human-led review?

The empirical results decisively answer all three in the affirmative.

\begin{enumerate}
    \item Accuracy – LLMs not only matched but significantly exceeded human performance in both invoice-level and line-item-level decision-making. Top-performing models achieved 92\% accuracy on invoice-level decisions and F-scores of up to 81\% for line-item classification—substantially higher than the 72\% and 43\%, respectively, achieved by the best-performing human reviewers. This demonstrates that LLMs can apply billing guidelines with greater consistency and precision, even when human discretion was factored into comparison.
    \item Efficiency – LLMs reviewed invoices 50x to 80x faster than human reviewers. While human invoice reviewers took 194 to 316 seconds per invoice, LLMs completed the same tasks in as little as 3.6 seconds. This highlights their ability to rapidly analyze, classify, and make billing decisions without loss of quality.
    \item Cost-Effectiveness – AI review costs were over 99\% lower than those of human invoice reviewers, with LLMs costing between \textbf{\$}0.003 and \$0.13 per invoice, versus \textbf{\$}4.17–\$4.47 per invoice for human reviewers. These findings support the conclusion that LLMs are not only more accurate and faster but also dramatically more scalable and financially viable.
\end{enumerate}

\begin{figure}
    \centering
    \includegraphics[width=1\linewidth]{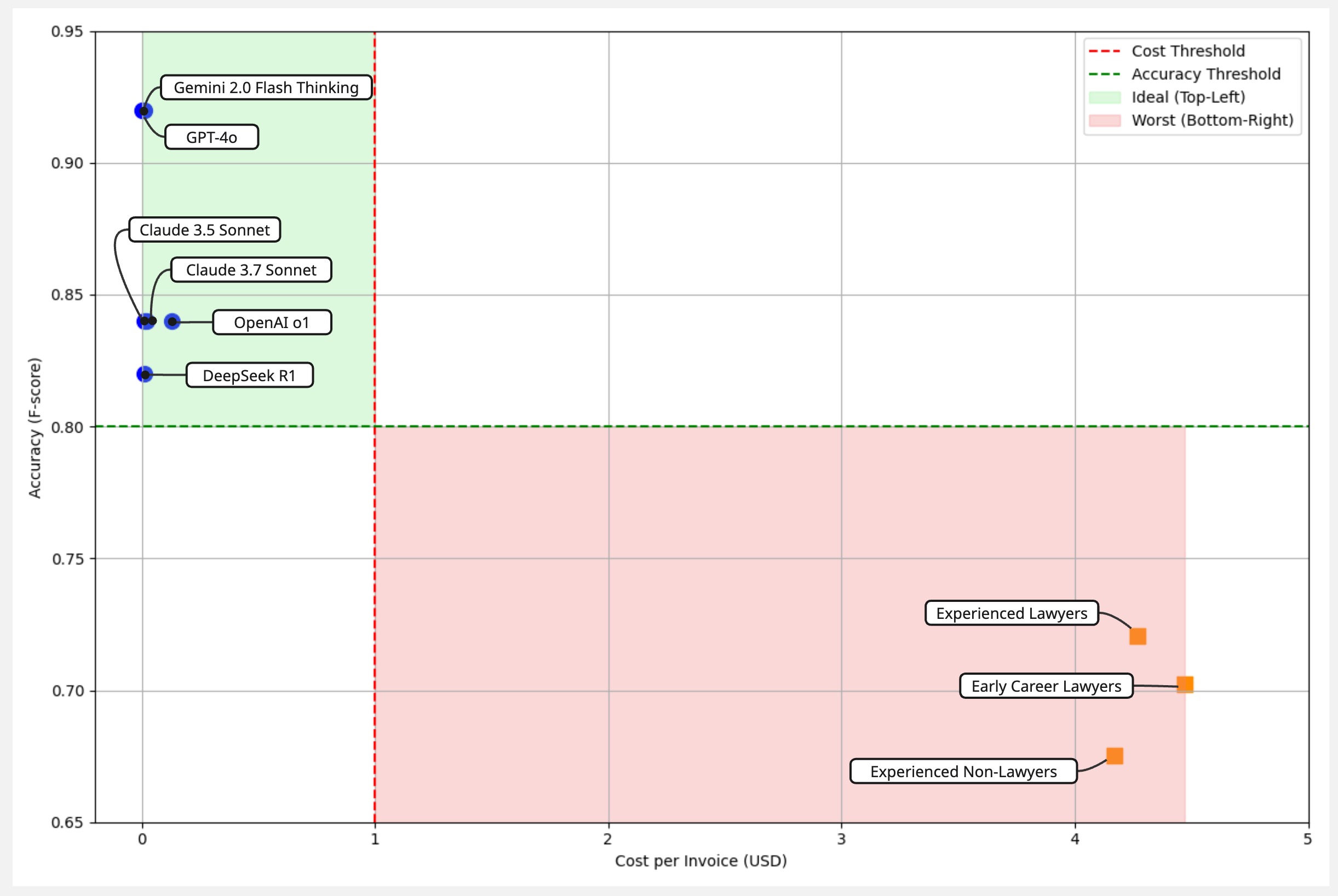}
    \caption{Invoice specific Cost vs. Accuracy Scatter Plot with Quadrant Analysis}
    \label{fig:invoice-cost-accuracy-scatter-plot}
\end{figure}

To further support the aforementioned observations, Figure~\ref{fig:invoice-cost-accuracy-scatter-plot} illustrates a clear efficiency advantage of AI systems over human professionals in legal invoice review, particularly within the top-left quadrant representing low cost and high accuracy. Notably, models like \textit{Gemini 2.0 Flash Thinking} and \textit{GPT-4o} achieve the highest accuracy at negligible cost, outperforming all Human Invoice Reviewers. All Human Invoice Reviewers fall outside the ideal quadrant, highlighting the disruptive potential of AI in domains traditionally reliant on expert judgement.

Together, these results present a compelling case for AI-driven legal invoice review. However, to fully consider the performance gap between LLMs and Human Invoice Reviewers, it is important to analyze why human reviewers struggled against the ground truth benchmark. While the Ground Truth team and Human Invoice Reviewers were presented with the same tasks, several factors likely contributed to the Human Invoice Reviewers lower accuracy:

\begin{enumerate}
    \item Cognitive load and rule complexity – Human invoice reviewers must recall and apply a wide range of billing rules, often under time constraints. Unlike LLMs, which can reference and enforce guidelines with high recall, human reviewers are susceptible to lapses in memory or inconsistent application of complex rule sets. This challenge is particularly pronounced in time-measured bulk tasks, such as this experiment, where cognitive fatigue can further affect performance.
    
    \item Ambiguity in plain language Guidelines\textbf{ }– The billing guidelines in this study were written in simplified plain English to improve readability. Although this makes them more accessible, it may also have introduced unintended ambiguity, leading to inconsistencies in human decision-making. LLMs, in contrast, systematically interpreted and enforced these guidelines with greater precision.
    
    \item Lack of context - Human invoice reviewers typically have a broader contextual understanding of legal matters, including the reasons behind certain line items and the history of client-firm interactions. In this study, their decision-making was restricted to the invoice and the billing guidelines, without access to additional matter context. This limitation likely contributed to their lower performance, as they were unable to apply the broader contextual reasoning they would normally use in practice.
\end{enumerate}

These insights suggest that, while AI-driven review may improve consistency and efficiency, the role of human expertise in providing context and handling ambiguity should not be overlooked. A key challenge for AI integration will be determining how to structure invoice review workflows to leverage AI’s strengths while mitigating its current limitations.

\subsection{Implications for the Legal Industry}

The findings suggest that LLMs can be a transformative tool in legal invoice review, offering organisations the ability to apply billing guidelines with greater consistency and efficiency. Historically, invoice review has been a labour-intensive process that relies on subjective human judgment. Our data indicates that AI can reduce this subjectivity and improve compliance enforcement.

The adoption of AI will likely reshape key players in the legal invoice review ecosystem:

\begin{itemize}
    \item Legal operations teams and third-party review services may shift from manual auditing to AI-driven oversight, focusing more on exception handling and compliance governance.
    
    \item Corporate legal departments could improve efficiency by enforcing billing policies with greater consistency while reducing the workload on legal teams.
    
    \item Law firms may face increased scrutiny, as AI-driven billing compliance could reduce discretionary billing practices and ensure more uniform enforcement of outside counsel guidelines.
    
\end{itemize}

Despite these advantages, adoption challenges remain. The transition to AI-driven review will not be determined solely by technical performance but also by broader industry dynamics, including regulatory considerations, client expectations, and resistance from stakeholders accustomed to manual review processes. Additionally, some aspects of invoice review — such as negotiations over disputed charges — may continue to benefit from human involvement. The key question is not whether AI will play a central role in legal invoice review, but how firms can best integrate it alongside human oversight to achieve optimal outcomes.

\subsection{Further Work }

While this study demonstrates that LLMs outperform human invoice reviewers in key areas, further research is needed to refine their application in real-world workflows. Several areas warrant additional investigation:

\begin{enumerate}
    \item Handling supporting documentation – Many billing entries rely on supplemental materials (e.g., timesheets, engagement letters). Future studies should examine how well LLMs can assess compliance when additional documentation is required.
    
    \item The role of professional judgment – While our study found that human discretion improved accuracy by 6\% on average, LLMs still outperformed human invoice reviewers overall. A key question for further research is whether this discretion provides meaningful value or whether it introduces variability that AI could mitigate.
    
    \item Hybrid AI-human workflows – Exploring structured workflows where human invoice reviewers intervene only in cases of ambiguity or dispute resolution could offer insights into how AI can complement rather than replace human expertise.
    
    \item Longitudinal impact – A longer-term study could examine whether AI-driven consistency leads to fewer invoice disputes over time or whether firms adjust their billing practices in response to AI-enforced scrutiny.
\end{enumerate}

Ultimately, while AI is poised to play a central role in legal invoice review, its optimal deployment will depend on balancing efficiency gains with the need for strategic human oversight.

\section{Conclusion}

This study challenges the traditional assumption that human expertise is indispensable in legal invoice review. Our findings indicate that LLMs not only match but surpass human invoice reviewers in key performance areas—accuracy, speed, and cost—suggesting that AI-driven processes could bring significant improvements to legal operations.

While widespread AI adoption may be gradual due to industry inertia and regulatory considerations, the efficiency and consistency benefits of LLMs are difficult to ignore. The real question is no longer whether AI can assist in invoice review, but how legal teams can integrate it effectively. Organizations that continue to rely on slower, costlier, and less consistent human invoice reviewers may find themselves at a competitive disadvantage as AI-driven alternatives become more prevalent.

Rather than viewing AI as a replacement for human invoice reviewers, legal teams should consider it as a strategic enhancement—one that can improve compliance, reduce costs, and allow professionals to focus on higher-value tasks. As the legal industry navigates this shift, the key challenge will be designing workflows that balance human expertise with AI’s capabilities in a way that maximizes efficiency without compromising judgment where it is truly needed.

\section*{Acknowledgements}

This research was conducted by Onit’s Artificial Intelligence Center of Excellence, with special mention to the principal investigators and supporters of the research:
Nick Whitehouse, Chief AI Officer (PI); Nicole Lincoln, Legal AI Engineer (Co-PI); Stephanie Yiu, Legal AI Engineer (Co-PI); Lizzie Catterson, Legal AI Engineer (Co-PI); Dr. Rivindu Perera, VP of AI and Data Science (Co-PI); Erin Cairney, Director of AI Trust and Integrity; Amruta Deshpane, AI Analyst; Dr. Jaron Mar, Principal AI Engineer; Eric Lim, Principal AI Engineer.

\bibliographystyle{abbrv}
\bibliography{bbgpt}

\appendix

\section{Appendix}

\subsection{Billing guidelines used in research}

\begin{enumerate}
    \item \textbf{Introduction}  
    You, as outside counsel, agree to follow these billing guidelines when providing legal services to our Company. These guidelines take effect on February 1, 2020 and replace any previous agreements. By submitting invoices, you agree to these terms. We reserve the right to modify these guidelines with prior written notice. Any invoices that do not comply with these guidelines may not be paid.
    
    Outside counsel may only be hired by an authorized representative of our legal department. A supervising attorney (\textit{Legal Contact}) will be assigned to each matter. All communication with Company personnel must go through the Legal Contact unless agreed otherwise.
    
    \item \textbf{Billing and Invoice Submission}  
    \begin{itemize}
        \item Invoices must be submitted electronically via our billing system on a monthly basis. Each invoice must cover only one matter and one calendar month.
        \item \textbf{Deadline}: Invoices must be submitted no later than the last day of the month following the month in which the work was performed.
        \item \textbf{Payment Terms}: Payment will be made within 60 days of the calendar month in which a properly submitted and accepted invoice is received.
        \item \textbf{Late Penalties}: If invoices are submitted late, a 5\% discount will be applied for every month the invoice is late, unless the invoice total is under \$500. Invoices submitted more than four months after the completion of work will not be paid.
    \end{itemize}
    
    \item \textbf{Line Item Details and Block Billing}  
    \begin{itemize}
        \item \textbf{Detailed Billing}: Block billing is not allowed. Each task must be listed separately, with the time spent on each clearly noted. Block billing will result in a 50\% reduction or rejection of the invoice.
        \item \textbf{Timekeeping}: Time must be recorded accurately by task, and each entry must describe the work performed and the timekeeper’s involvement.
    \end{itemize}
    
    \item \textbf{Prohibited Fees and Expenses}  
    We will not pay for:
    \begin{itemize}
        \item Administrative, clerical, or secretarial work (e.g., word processing, proofreading).
        \item Time spent getting 'up to speed' due to staff turnover or absences.
        \item Invoice preparation, billing inquiries, or disputes.
        \item Professional development or training time for attorneys.
        \item Document scanning, copying, or communication charges (e.g., phone, fax).
        \item Overhead expenses (e.g., meals, local transportation, conference room rentals).
    \end{itemize}
    
    \item \textbf{Travel Expense Policy}  
    We will reimburse reasonable travel expenses according to the following guidelines:
    \begin{itemize}
        \item \textbf{Air Travel}: We reimburse economy class only, but flexible travel times and bookings are encouraged.
        \item \textbf{Hotel}: Reimbursement up to \$300 per night.
        \item \textbf{Meals}: Meal expenses are reimbursed up to \$75 per day.
        \item \textbf{Ground Transportation}: Rental cars will be reimbursed up to \$50 per day. Limousine services are not covered unless they are cheaper than taxi fares.
    \end{itemize}
    
    \item \textbf{Budgeting and Forecasts}  
    Within 10 days of opening a new matter, outside counsel must submit a detailed budget. This budget should be updated monthly and must reflect a rolling 12-month forecast.
    
    \item \textbf{Staffing}  
    \begin{itemize}
        \item \textbf{General Staffing Guidelines}: No more than 8 billable hours per day should be billed by a timekeeper. We will not pay for time spent by staff learning the matter due to turnover or other reasons.
        \item \textbf{Expertise}: Team members must have appropriate expertise for their role, and first-year associates must not be assigned to our matters.
        \item \textbf{Diversity}: We encourage outside counsel to promote diversity in staffing.
    \end{itemize}
    
    \item \textbf{Third-Party Services}  
    Third-party vendors may be engaged, provided the costs do not exceed \$5,000 per vendor. Notify us if costs are expected to exceed this threshold.
    
    \item \textbf{Performance Reviews}  
    We will conduct regular performance reviews, assessing compliance with these guidelines and cost efficiency.
    
    \item \textbf{Ownership and Disposition of Documents}  
    All work created for our Company is our property. Upon request, you will promptly provide copies of all legal documents and work product.
    
    \item \textbf{Media and Public Inquiries}  
    Outside counsel may not use our Company name, logos, or trademarks in any marketing materials without prior written consent.
    
    \item \textbf{Litigation and Patent Matter Requirements}  
    \begin{itemize}
        \item \textbf{Litigation}: eDiscovery must follow our protocol. Document production should be managed cost-effectively.
        \item \textbf{Patent}: Patent filings must be timely, and staffing must be efficient.
        \item \textbf{General}: Monthly reporting and cost tracking are required.
    \end{itemize}
    
    \item \textbf{Miscellaneous Provisions}  
    \begin{itemize}
        \item \textbf{Metrics and Reporting}: Quarterly reports must be provided, including invoiced amounts and staffing details.
        \item \textbf{Conflicts of Interest}: Outside counsel must notify us of any potential conflicts of interest before beginning work.
    \end{itemize}
    
    \item \textbf{Dispute Resolution}  
    \begin{itemize}
        \item \textbf{Escalation of Disputes}: Disputes must first be escalated internally.
        \item \textbf{Mediation}: If unresolved, disputes will be submitted to mediation, with costs shared equally.
    \end{itemize}
    
    \item \textbf{Termination}  
    Either party may terminate the agreement with 30 days' notice. Upon termination, all work and documents must be returned or destroyed within 15 days.
\end{enumerate}

\end{document}